%% file: paper.tex
\documentclass[]{bytedance_seed}
\pdfoutput=1

\usepackage[toc,page,header]{appendix}

\usepackage{minitoc}
\usepackage{cleveref} 
\usepackage{subcaption}
\usepackage{booktabs}
\usepackage{tabularx}

\usepackage{lipsum} 

\usepackage{amsmath,amssymb}

\input{macro}
\title{Seed Diffusion: A Large-Scale Diffusion Language Model with High-Speed Inference}

\affiliation[1]{ByteDance Seed\quad $^2$Institute for AI Industry Research (AIR), Tsinghua University \newline $^3$SIA-Lab of Tsinghua AIR and ByteDance Seed}

\abstract{\input{input/000abstract}}

\date{\today}
\correspondence{\email{zhouhao@air.tsinghua.edu.cn}, \email{zhangzheng.jacob@bytedance.com}}

\checkdata[Project Page]{\url{https://seed.bytedance.com/seed_diffusion}}

\begin{document}

\newtheorem{theorem}{Theorem}[section]
\newtheorem{proposition}[theorem]{Proposition}
\newtheorem{lemma}[theorem]{Lemma}
\newtheorem{corollary}[theorem]{Corollary}
\newtheorem{definition}[theorem]{Definition}
\newtheorem{assumption}[theorem]{Assumption}
\newtheorem{remark}[theorem]{Remark}

\maketitle

\input{input/010intro}
\input{input/020related}
\input{input/030method}

\input{input/031method}

\input{input/032method}
\input{input/033method}
\input{input/040experiments}
\input{input/050conclusion}

\clearpage

\bibliographystyle{unsrt}
\bibliography{main}

\clearpage

\end{document}

%% file: macro.tex
\usepackage{natbib}
\usepackage{latexsym}

\usepackage{url}
\usepackage{amssymb}
\usepackage[utf8]{inputenc}
\usepackage{microtype}
\usepackage{booktabs}
\usepackage{pifont} 
\usepackage{multirow}
\usepackage{makecell}
\usepackage{paralist}
\usepackage{xspace}
\usepackage{color}
\usepackage{xcolor}
\usepackage{colortbl}
\usepackage{adjustbox}
\usepackage{hyperref} 
\usepackage[edges]{forest}
\usepackage{tikz} 
\usepackage{caption}
\usepackage{amsfonts}

\hypersetup{
    colorlinks,
    linkcolor={blue!80!black},
    citecolor={blue!80!black},
}
\tikzset{
    root/.style =             {align=center, text width=1cm, rounded corners=3pt, line width=0.3mm, fill=gray!10, draw=gray!80, font=\small},
    demographic/.style =         {align=center, text width=1.8cm, rounded corners=3pt, line width=0.3mm, fill=blue!10, draw=blue!80, font=\footnotesize},
    demographic_work/.style =    {align=center, text width=10cm, rounded corners=3pt, line width=0.3mm, fill=blue!10, draw=blue!0, font=\footnotesize},
    character/.style =         {align=center, text width=1.8cm, rounded corners=3pt, line width=0.3mm, fill=red!10, draw=red!80, font=\footnotesize},
    character_work/.style =    {align=center, text width=10cm, rounded corners=3pt, line width=0.3mm, fill=red!10, draw=red!0, font=\footnotesize},
    personalization/.style =           {align=center, text width=1.8cm, rounded corners=3pt, line width=0.3mm, fill=cyan!10, draw=cyan!80, font=\footnotesize},
    personalization_work/.style =      {align=center, text width=10cm, rounded corners=3pt, line width=0.3mm, fill=cyan!10, draw=cyan!0, font=\footnotesize},
    risk/.style =         {align=center, text width=1.8cm, rounded corners=3pt, line width=0.3mm, fill=orange!10, draw=orange!80, font=\footnotesize},
    risk_work/.style =    {align=center, text width=10cm, rounded corners=3pt, line width=0.3mm, fill=orange!10, draw=orange!0, font=\footnotesize},
}

\usepackage{CJK}

%% file: input/000abstract.tex
We present \textbf{Seed Diffusion Preview}, a large-scale language model based on discrete-state diffusion, offering remarkably fast inference speed.
Thanks to non-sequential, parallel generation, discrete diffusion models provide a notable speedup to mitigate the inherent latency of token-by-token decoding, as demonstrated recently (e.g., Mercury Coder~\cite{khanna2025mercury}, Gemini Diffusion~\cite{gdm2025geminidiffusion}).  
Seed Diffusion Preview achieves an inference speed of \textbf{2,146 token/s} over H20 GPUs while maintaining competitive performance across a sweep of standard code evaluation benchmarks, significantly faster than contemporary Mercury and Gemini, establishing new state of the art on the speed-quality Pareto frontier for code models. 
Demo is available at \url{https://studio.seed.ai/exp/seed_diffusion/}. 

%% file: input/010intro.tex
\vspace{-20px}
\begin{figure}[h]
    \centering
    \includegraphics[width=\linewidth]{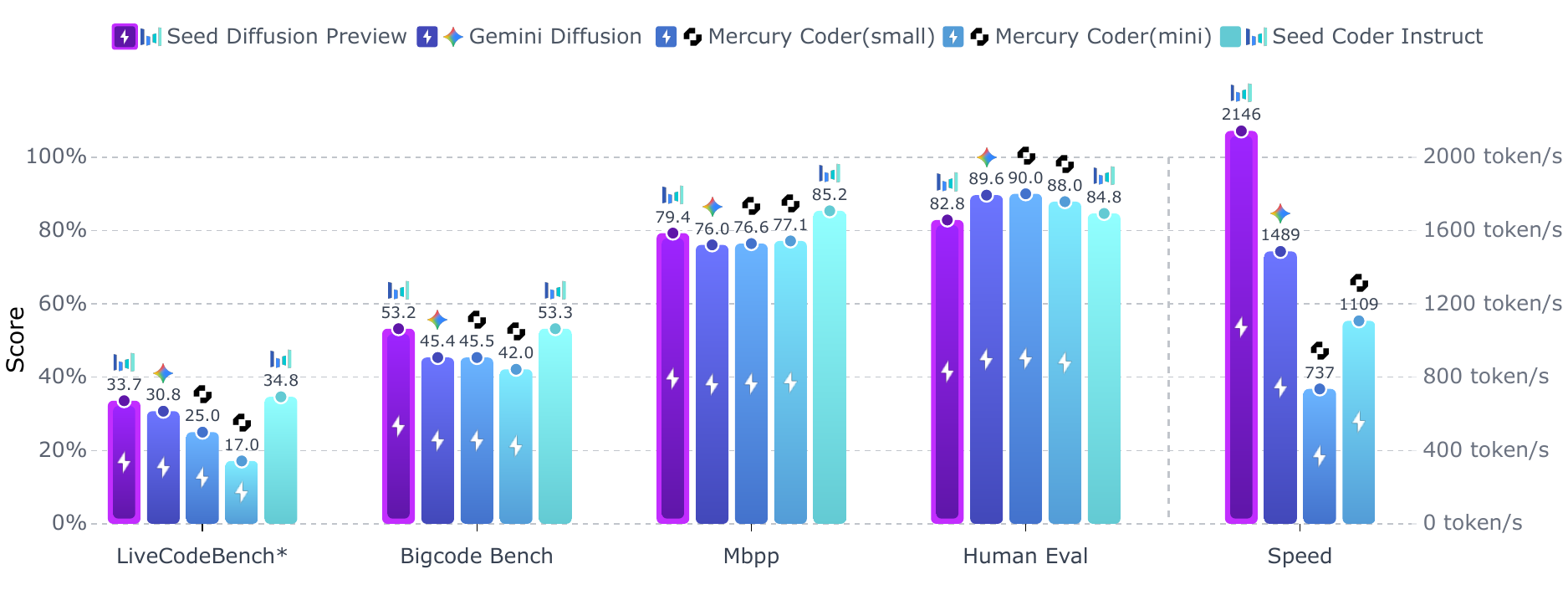}
    \caption{Seed Diffusion's inference speed is measured over H20 GPUs across eight open code benchmarks. Direct comparison with baselines is challenging due to differing test conditions: Mercury Coder was evaluated on a proprietary dataset with H100s, while Gemini Diffusion's speed was averaged over a mixed-task benchmark using unknown hardware. Furthermore, reported speeds on these benchmarks can benefit from format-constraining system prompts. LiveCodeBench results are specifically on the 1055 problems from v1-v6 for the unknown baselines' protocol.}
    \label{fig:front}
\end{figure}

\newpage

\section{Introduction}
Diffusion models~\citep{ho2020denoising,song2019generative,sohl2015deep} learn to reverse a process that incrementally corrupts data with noise, effectively decomposing a complex distribution into a hierarchy of simplified representations. This coarse-to-fine generative approach has proven remarkably successful across a wide range of applications, including image and video synthesis~\cite{ho2022video} as well as solving complex challenges in natural sciences~\citep{varadi2024alphafold}. 

However, translating this success to the discrete domain of natural language presents critical challenges. The primary difficulty stems from the fact that standard diffusion process is naturally defined over continuous state spaces, thus not directly applicable to discrete domains such as natural language. To bridge this gap, many efforts have focused on novel adaptations, ranging from projecting discrete tokens into a continuous latent space (e.g., embeddings or a simplex) where diffusion can be applied \citep{graves2023bayesian,dieleman2022continuous,gulrajani2024likelihood}, to constructing the diffusion process directly over discrete-state space by defining explicit state transition matrices \citep{austin2021structured,sahoo2024simple,shi2024simplified,ou2024your}. Recent discrete-state approaches have demonstrated scalability and effectiveness with advanced architectures and training recipes~\cite{nie2025large}.

Despite impressive progress, real-world deployment of discrete diffusion models for language is still hampered by two key challenges:

\begin{itemize}
        \item \textbf{Inductive bias on token–order modeling.}  
          The usage of discrete diffusion for modeling and generating tokens in arbitrary orders is theoretically powerful and appealing~\cite{shi2024simplified,ou2024your}; however, natural language is overwhelmingly processed in a sequential order.  
          A purely random-order learning signal can in consequence be inefficient, or even detrimental for language modeling, dampening model performance.

    \item \textbf{Inference inefficiency.}  
          Although diffusion models are non-autoregressive, their iterative step-sensitive denoising procedure introduces severe latency, which undermines their major advantage over traditional autoregressive models, acting as a cumbersome bottleneck in practice.
\end{itemize}

In this work, we introduce Seed Diffusion Preview, a code-focused language model designed to achieve an elegant balance between speed and quality. Tackling these challenges directly, our model achieves a remarkable speed of 2146 tokens/second on H20 GPUs while maintaining competitive performance against similarly-sized standard language models across a diverse set of code evaluation benchmarks, establishing new state of the art on the speed-quality Pareto frontier. 

%% file: input/020related.tex
\section{Related Work}

\textbf{Non-autoregressive (NAR)} models have long been considered an alternative to sequential decoding, valued for their potential of parallel inference. In the pre-LLM era, many early NAR methods demonstrated strong performance on specific tasks such as machine translation~\cite{qian2020glancing,huang2022directed,ghazvininejad2019mask}. However, these approaches often lacked a rigorous theoretical foundation for density estimation, which limited their viability as general-purpose probabilistic language models.

\textbf{Discrete diffusion} models~\cite{austin2021structured, lou2023discrete, sahoo2024simple,ou2024your,shi2024simplified} have emerged to close this gap. By optimizing the Evidence Lower Bound (ELBO), they provide a principled probabilistic framework for language modeling. The recent success of large-scale systems such as Mercury Coder~\cite{khanna2025mercury} and Gemini Diffusion~\cite{gdm2025geminidiffusion} is particularly notable. These models show that it is possible to narrow the quality gap with autoregressive systems while offering substantial speedup, thereby challenging the conventional wisdom on "quality-speed trade-off",  raising new interest in NAR in the modern LLM era. 

%% file: input/030method.tex
\section{Seed Diffusion}

As the first experimental model in our Seed Diffusion series, Seed Diffusion Preview is specifically focused on code generation, thus adopting the data pipelines (\textbf{code/code-related data only}) and processing methodology of the open-sourced Seed Coder project~\cite{zhang2025seed}. The architecture is a standard dense Transformer, and we intentionally omit complex components such as LongCoT reasoning in this initial version to first establish a strong and efficient performance baseline. This section introduces its key components and training  strategies.

\subsection{TSC: A Two-Stage Curriculum for Robust Diffusion Training}
\label{sec:diff_training}
The first stage is \textit{scaled diffusion training}. Seed Diffusion Preview is a discrete-state diffusion language model trained with two types of forward corruption process. 
Given an initial data sequence $\mathbf{x}_0 \sim p_{\text{data}}$ and continuous timestep settings where $t\sim[0,1]$, the forward process implied by the marginal $q(\mathbf{x}_t|\mathbf{x}_0)$ is defined as follows:

\paragraph{\textbf{Mask-based Forward Process}}
For the first 80\% diffusion training steps, we use a standard mask-based corruption process~\cite{sahoo2024simple,nie2025large}. This process gradually replaces tokens in the original sequence $\mathbf{x}_0$ with a special \texttt{[MASK]} token($\mathbf{m}$). The corrupted sequence $\mathbf{x}_t$ is sampled from the conditional distribution $q(\mathbf{x}_t|\mathbf{x}_0)$ where each token is treated independently:
\begin{equation}
    q_{\text{mask}}(\mathbf{x}_t|\mathbf{x}_0) = \prod_{i=1}^{|\mathbf{x}_0|} q_{\text{mask}}(\mathbf{x}_t[i] | \mathbf{x}_0[i])
\end{equation}
The probability of a token remaining unchanged or being masked is determined by a noise schedule $\gamma_t$. For any position $i$, the marginal probability is:
\begin{equation}
    q(\mathbf{x}_t[i] = c | \mathbf{x}_0[i]) = 
    \begin{cases}
        1- \gamma_t & \text{if } c = \mathbf{x}_0[i] \\
        \gamma_t & \text{if } c = \mathbf{m}
    \end{cases}
\end{equation}
$\gamma_t$ refers to the noise schedule function designed to be monotonically increased~\cite{qian2020glancing,qian-etal-2024-diffusion}.
\paragraph{\textbf{Edit-based Forward Process}}
For the last 20\% diffusion training steps, we add an extra edit-based corruption process as augmentation to improve calibration and eliminate unexpected behavior such as repetitions in the sampling process. Similar to mask-based approaches, we control a designed signal-to-noise ratio based on \textbf{Levenshtein distance}, $d_{\text{Lev}}(\mathbf{x}_a, \mathbf{x}_b)$, which measures the minimum number of token-level edits required to change one sequence into another~(refer to \cite{qian2020glancing} for more insights).  
The forward process then samples a corrupted sequence based on a predefined edit operation set (e.g., deletions, insertions, and substitutions) and defines the total edit-operation number as $k_t$ to approximately control langevin distance. The $q_\text{edit}(\mathbf{x}_t | \mathbf{x}_0)$ is implicitly defined by: 
\begin{align}
    \mathbf{z}_0 = \mathbf{x}_0;  \quad \mathbf{z}_j = o_j(\mathbf{z}_{j-1}) \quad \text{for } j = 1, \dots, k_t, o \in \mathcal{O}; \quad \mathbf{x}_t = \mathbf{z}_{k_t}
\end{align}
$\mathcal{O}$ is a predefined operations set. Although applying $k_t$ edits does not guarantee that the final Levenshtein distance $L(\mathbf{x}_0, \mathbf{x}_t)$ is exactly $k_t$ (e.g., an insertion followed by a deletion can cancel out), it provides a tractable and scalable method to control the corruption level. The target number of edits $k_t$ is scheduled as:
\begin{align}
    k_t = \lfloor |\mathbf{x}_0| \cdot (\alpha_t) \rfloor
\end{align}
Here $\alpha_t$ denotes the scheduler for the approximate signal-to-noise ratio. As an auxiliary objective, we make $\alpha_t$ lie in the range $[0,0.1]$ to maintain the density estimation ability of mask-based forward process.

\textbf{Overall Learning Objective}  
The reverse process is  parameterized as $p_\theta(\mathbf{x}_s | \mathbf{x}_t)$. We set the predicted probability over the mask token always as 0~\cite{sahoo2024simple} by adding $-\inf$ to the corresponding logits. With this formulation, the mask-based forward process implies an analytical posterior, hence a tractable and simple formulation ELBO: 
\begin{align}
\label{eq:elbo}
    L_{\text{ELBO}} =  \underbrace{-\log p_\theta\left(\mathbf{x}_0 \mid \mathbf{x}_{0}\right)}_{\text{Recontruct Loss}} -\mathbb{E}_{q_{\text{mask}}, t}\left[ \frac{\gamma_t^{'}}{\gamma_t} \sum_{i=1}^{|\mathbf{x}_0|} \mathbf{1}\left[\mathbf{x}_t[i]=\mathbf{m}\right] \log p_\theta\left(\mathbf{x}_0[i] \mid \mathbf{x}_t[i]\right)\right]
\end{align}
Our learning objective derives from ELBO in Eq.~\ref{eq:elbo} by substituting the reconstruct loss with a denoised loss based on the edit-based forward process $q_\text{edit}$ as:
\begin{align}
\label{eq:elbo}
    L_{\text{diff}}{(\theta)} =  -\mathbb{E}_{q_{\text{edit}},t} \log p_\theta(\mathbf{x}_0 | \mathbf{x}_t)  -\mathbb{E}_{q_{\text{mask}}, t}\left[ \frac{\gamma_t^{'}}{\gamma_t} \sum_{i=1}^{|\mathbf{x}_0|} \mathbf{1}\left[\mathbf{x}_t[i]=\mathbf{m}\right] \log p_\theta\left(\mathbf{x}_0[i] \mid \mathbf{x}_t[i]\right)\right]
\end{align}

\begin{remark}
Unlike some prior work~\cite{sahoo2024simple,shi2024simplified}, we do not employ the strategy of "Carry Over Unmasking", \emph{i.e.} directly copying unmasked input tokens to the output, despite  potential benefits to perplexity. While a purely mask-based diffusion process offers a low-variance training objective (each position is either the ground truth or a [MASK] token), it introduces a detrimental inductive bias. Such a model learns a spurious correlation that unmasked tokens are always correct, leading to overconfidence and unable to perform self-correction during inference. To mitigate this, our edit-based augmentation forces the model to re-evaluate all tokens, including those unmasked.
\end{remark}

%






%% file: input/031method.tex
\subsection{Tailoring the Trajectory Space of Diffusion}
\label{sec:anyorder}
An important perspective to understand the essentials of mask-based diffusion models is its equivalence to any-order autoregressive modeling, which has been revealed and illustrated by \cite{hoogeboom2021autoregressive}. This perspective builds the correlation between ELBO and an expected log-likelihood of any-order autoregressive models as:

\begin{equation}
    \mathcal{J}(\theta) = -\mathbb{E}_{\pi \sim U(\mathcal{S}_d)} \left[ \sum_{r=1}^{d=|\mathbf{x}|} \log p_\theta(\mathbf{x}_{\pi(r)} \mid \mathbf{x}_{\pi(<r)}) \right]
\end{equation}

Here $S_{d}$ stands for the symmetric group of all possible permutations $\pi$ over $\{0,1, \ldots, d-1\}$. $p_\theta(\mathbf{x}_{\pi(r)} \mid \mathbf{x}_{\pi(<r)})$ models the conditional probability of $\mathbf{x}_{\pi(r)}$ given all preceding tokens in the permutations $\pi$ denoted as  $\mathbf{x}_{\pi(<r)}$. 
Now recall that with the transition probability $q_{\text{mask}}(\mathbf{x}_t|\mathbf{x}_s),t>s$ denotes as:
\begin{align}
q_{\text{mask}}(\mathbf{x}_t[i] \mid \mathbf{x}_s[i]) =
\begin{cases}
\frac{1-\gamma_t}{1-\gamma_s} & \text{if } \mathbf{x}_t[i] = \mathbf{x}_s[i] \neq \mathbf{m} \\
\frac{\gamma_t - \gamma_s}{1-\gamma_s} & \text{if } \mathbf{x}_t[i] = \mathbf{m} \text{ and } \mathbf{x}_s[i] \neq \mathbf{m} \\
1 & \text{if } \mathbf{x}_t[i] = \mathbf{x}_s[i] = \mathbf{m} 
\end{cases}
\end{align}
Intuitively, we interpret any trajectory $\tau$ with $K$ elements $\tau = \{\mathbf{x}_0 \rightarrow \cdots\rightarrow\mathbf{x}_i\rightarrow\mathbf{x}_K\}$ ($\mathbf{x}_K$ are all mask tokens) obtained based on the above transition conditional probability as an order of autoregressive decomposition.

Mask-based diffusion training presents a more complex learning problem than standard left-to-right autoregressive (AR) training. 
By design, diffusion models must learn from all possible generation orders, including many that are redundant, detrimental, or misaligned with the natural structure of language~\cite{miao2020do}. Consequently, diffusion-trained language models lag significantly behind their AR counterparts, even on code data that lacks a strong left-to-right prior. This persistent gap presents a fundamental challenge for the diffusion approach.

We propose a constrained-order diffusion training process after the two-stage diffusion
learning. This procedure involves creating a distilled dataset of optimal generation trajectories. Specifically, for any given sample, a candidate pool of trajectories is generated at scale using the pre-trained diffusion model. A selection criterion based on maximizing the Evidence Lower Bound (ELBO) is applied to filter this pool, and the resulting high-quality trajectories are used to fine-tune the model.
With the high-quality synthesized trajectories as $\mathrm{T}$, the constrained-order training takes the form of the following:
\begin{align}
    L_{\text{c}}(\theta) =\mathbb{E}_{\tau\sim U(\mathrm{T}),(\mathbf{x}_i,\mathbf{x}_0) \in \tau} - \lambda(\mathbf{x}_i) \log p_\theta (\mathbf{x}_0 | f(\mathbf{x}_i)) 
\end{align}
$\lambda(\mathbf{x}_i)$ is the weight for balancing loss toward different noise levels, and $f$ is an augmentation function similar to $q_\text{edit}$ in Section.~\ref{sec:diff_training}. 









%% file: input/032method.tex
\subsection{On-policy Diffusion Learning}
Although in theory discrete diffusion models should offer the advantage of parallel decoding, in practice realizing this potential is challenging. A single parallel inference step is computationally expensive, meaning a large number of tokens must be generated simultaneously to amortize this overhead in order to achieve actual inference efficiency gains. Reducing the total number of generation steps for better efficiency, however, can result in severe degradation in performance, especially in mask-based approaches~\cite{nie2025large}.


To fully unlock the parallel power, we propose a simple yet effective on-policy learning paradigm: for the reverse process parameterized with $\theta$, we optimize the objective as: 

\begin{align}
\label{eq:on-policy}
\mathbb{E}_{\substack{
    \text{prompt} \sim p_{\text{data}} \\
    \tau \sim p_\theta(\cdot|\text{prompt})
}}
\left[ |\tau| - V(\tau[0]) \right]
\end{align}

Here $\tau = \{\tau[K], \cdots,\tau[i],\cdots,\tau[0]\}$ is the sampled trajectory of the reverse process with a strategic sampling strategy, and the model $\theta$ is conditioned on the given prompts. The trajectory starts from the sequence with all mask tokens, and the final generated samples are $\tau[0]$.  $|\tau|$ denotes the sample steps and $V(\cdot)$ represents a model-based verifier that ensures the sampling process always converges to a reasonable/correct sample.  The verifier-related term ($V(\tau[0])$) in Equation.~\ref{eq:on-policy} can be optimized with the log-derivative trick~\cite{mohamed2020monte}. We observed that directly minimizing trajectory length led to unstable training dynamics. Therefore, we optimize a progressive surrogate loss based on the fact that
\begin{align}
    |\tau| \propto \mathbb{E}_{i,j \in \{0,\cdots,K\}} \frac{1}{d_{\text{Lev}}(\tau[i],\tau[j])}
\end{align}

The speed-up dynamics during on-policy training is illustrated in Figure~\ref{fig:speedup_training_step}. Interestingly, this procedure has an effect analogous to mode filtering, a technique previously explored in the non-autoregressive (NAT) text generation literature~\cite{qian2020glancing,gu2017non}. 




%% file: input/033method.tex
\begin{figure}[!htbp]
    \centering
    \begin{subfigure}{0.48\textwidth}
        \includegraphics[width=\textwidth]{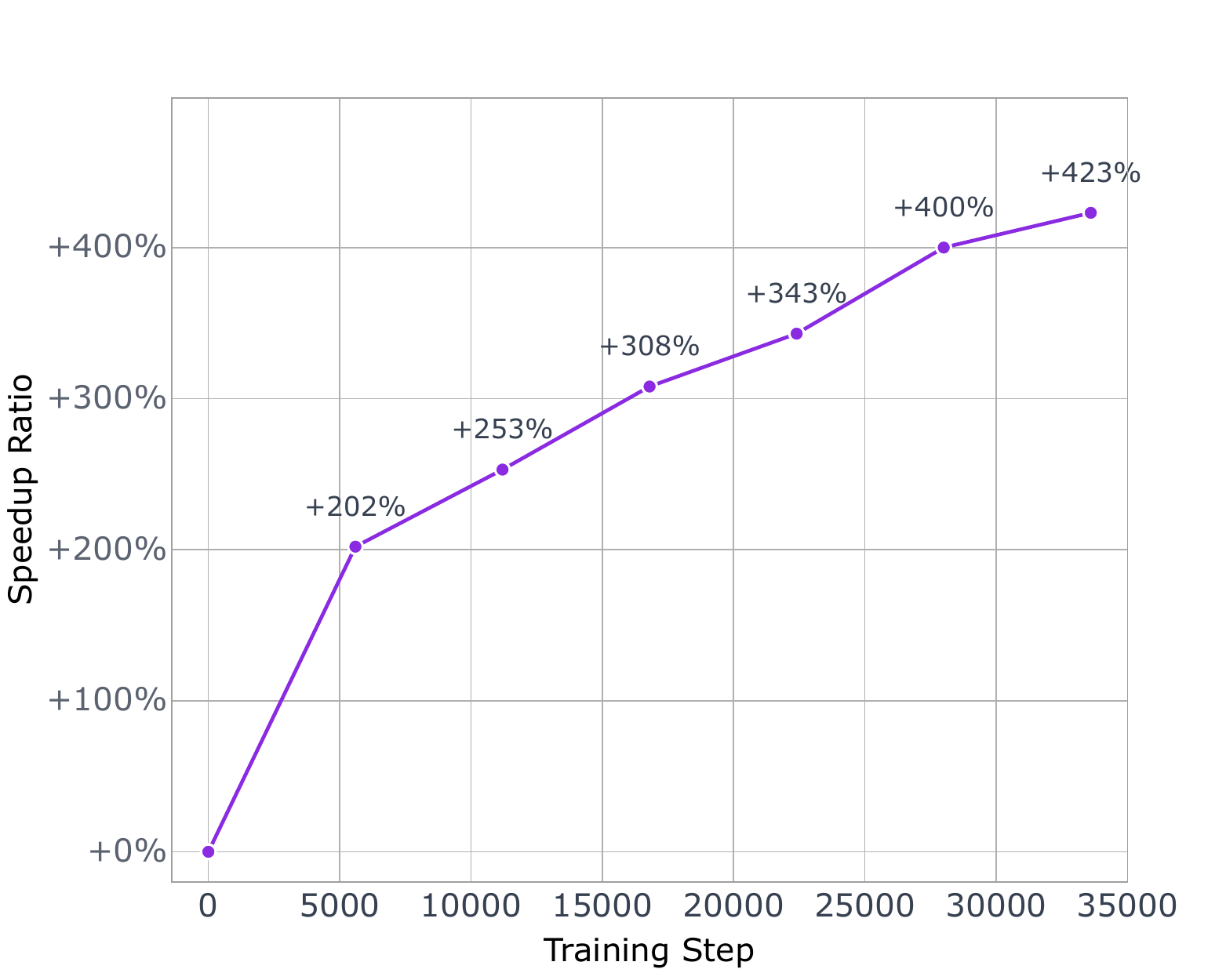}
        \caption{The changes of speedup ratio estimated by sampling with a certain block size $b$ during the on-policy training.}
        \label{fig:speedup_training_step}
    \end{subfigure}
    \hfill
    \begin{subfigure}{0.48\textwidth}
        \includegraphics[width=\textwidth]{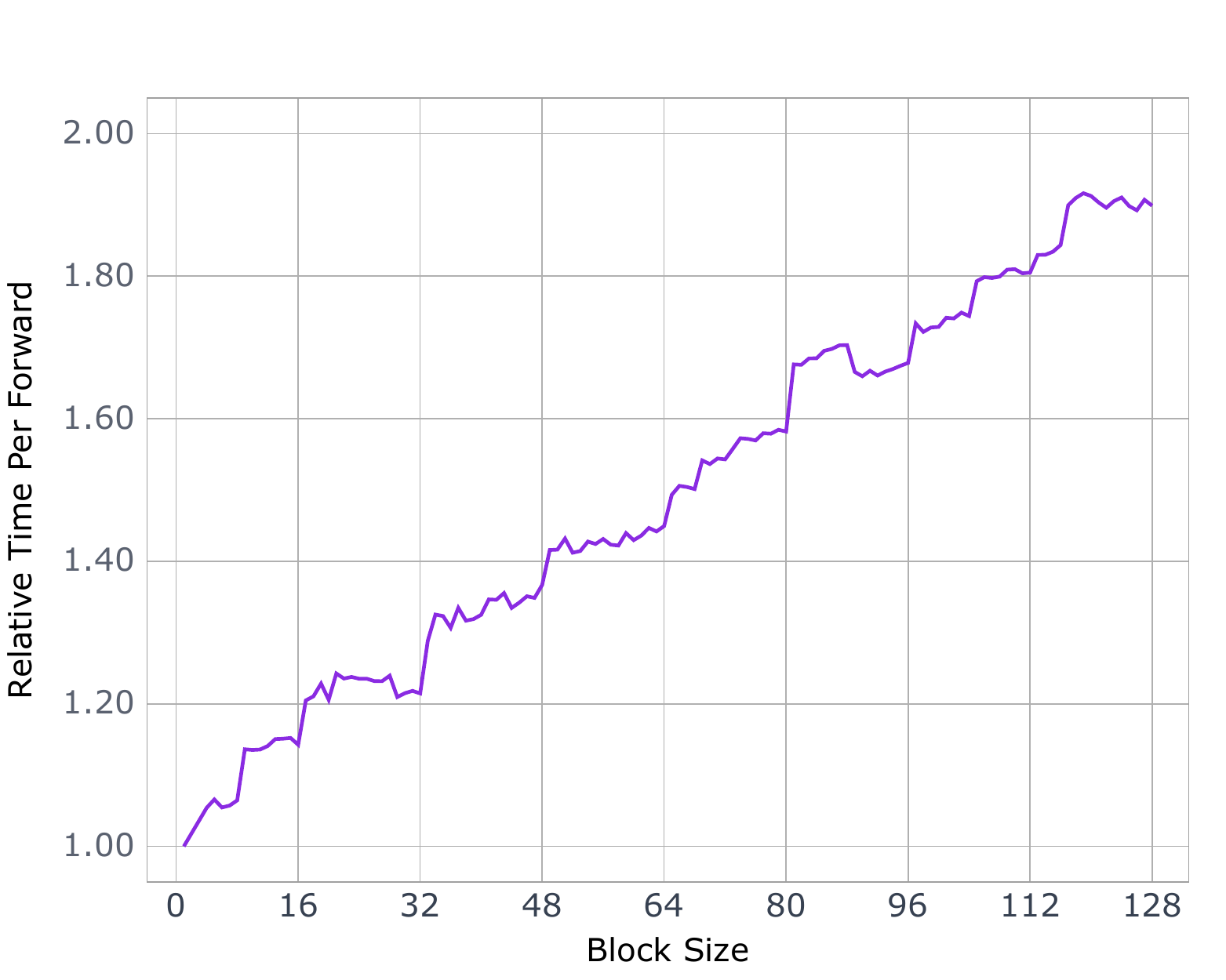}
        \caption{The relative forward time($\frac{T(B=b)}{T(B=1)}$) changes over the different block size($b$)}
        \label{fig:speedup_block_size}
    \end{subfigure}
    \hfill
    \caption{On-policy Training Dynamics $\And$ Block-wise Inference Time}
    \label{fig:main_text_fig}
\end{figure}
\subsection{Inference and Infrastructure}

To balance computation and latency, we employ a block-level parallel diffusion sampling scheme that maintains a causal ordering between blocks. For generating tokens in the $n$-th block  ($B_n$), the reverse process is expressed as $p_\theta(\mathbf{x}_t | \mathbf{x}_s,\mathbf{x}^{B_0,\cdots,B_n})$. Here $\mathbf{x}^{B_0,\cdots,B_n}$ denotes the previously generated block of tokens with $\mathbf{x}^{B_0} = \emptyset$.  
This semi-autoregressive (semi-AR) process is a well-established technique for balancing generation quality and efficiency~\cite{nie2025large,arriola2025block,hu2024acdit}. 
We avoid block-specific training\cite{arriola2025block,hu2024acdit} to retain  flexibility for arbitrary block partitioning during inference. We use KV-caching for previously generated blocks to condition subsequent ones. Although this risks potentially introducing potential bias, we empirically observe no significant degradation in generation quality. This robustness is probably due to the distillation of constrained-order trajectories as introduced in Section~\ref {sec:anyorder}. 

Beyond algorithmic design, our work employs holistic system optimization to support block-level inference efficiently. Specifically, we leverage our internal infrastructure framework, featuring specialized optimizations for diffusion sampling, to accelerate generation. The impact on performance across different block sizes is detailed in Figure~\ref{fig:speedup_block_size}. This analysis informs our selection of the optimal block size, determined by the trade-off between the latency of a single forward pass and the corresponding token generation rate.




%% file: input/040experiments.tex
\section{Experiments}

We benchmark the performance and decoding speed of Seed Diffusion across a range of code-related tasks. Our evaluation protocols and primary baselines are adapted from \cite{zhang2025seed}. To provide a  comprehensive comparison, we also include state-of-the-art Diffusion Language Models for experiments:  Mercury~\cite{khanna2025mercury} and Gemini-Diffusion~\cite{gdm2025geminidiffusion}.   







\subsection{Benchmarks}

\begin{table}[!t]
    \centering
    \caption{Performance on Aider ("whole" format) and CanItEdit.}
    \label{tab:performance}
    \begin{tabular}{llcc}
        \toprule
        \textbf{Model} & \textbf{Size} & \textbf{Aider} & \textbf{CanItEdit} \\
        & & \textbf{tries=2} & \textbf{pass@1} \\
        \midrule
        \multicolumn{4}{c}{\textbf{$\leq$15B Models}} \\
        \midrule
        CodeLlama-7B-Instruct & 7B & 1.5 & 25.7 \\
        DeepSeek-Coder-6.7B-Instruct & 6.7B & 44.4 & 36.9 \\
        CodeQwen1.5-7B-Chat & 7B & 38.3 & 34.8 \\
        Yi-Coder-9B-Chat & 9B & 54.1 & 50.5 \\
        Qwen2.5-Coder-14B-Instruct & 14B & \textbf{69.2} & \textbf{52.9} \\
        StarCoder2-15B-Instruct & 15B & 38.2 & 31.4 \\
        Llama-3.1-8B-Instruct & 8B & 33.1 & 39.5 \\
        OpenCoder-8B-Instruct & 8B & 30.8 & 39.0 \\
        Qwen2.5-Coder-7B-Instruct & 7B & \textbf{57.9} & 49.5 \\
        Qwen3-8B & 8B & 55.6 & 45.7 \\
        Seed-Coder-8B-Instruct & 8B & 57.1 & 50.5 \\
        \textbf{Seed-Diffusion-Preview}(0705) &  - & 44.4 & \textbf{54.3} \\
        \midrule
        \multicolumn{4}{c}{\textbf{15B+ Models}} \\
        \midrule
        Codestral-22B & 22B & 51.1 & 52.4 \\
        CodeLlama-70B-Instruct & 70B & 15.0 & 40.5 \\
        DeepSeek-Coder-33B-Instruct & 33B & 54.5 & 46.2 \\
        DeepSeek-Coder-V2-Lite-Instruct & 2.4B/16B & 52.6 & 45.2 \\
        \bottomrule
    \end{tabular}
\end{table}

\begin{table}[htbp]
    \centering
    \caption{Performance on MBXP.}
    \label{tab:mbxp_performance}
    \begin{adjustbox}{max width=\textwidth}
    \begin{tabular}{lccccccccccccccc}
        \toprule
        \textbf{Model} & \textbf{Size} & \textbf{Python} & \textbf{Java} & \textbf{C++} & \textbf{C\#} & \textbf{TS} & \textbf{JS} & \textbf{PHP} & \textbf{Go} & \textbf{Kotlin} & \textbf{Perl} & \textbf{Ruby} & \textbf{Scala} & \textbf{Swift} & \textbf{Average} \\
        \midrule
        \multicolumn{16}{c}{\textbf{$\leq$15B Models}} \\
        \midrule
        CodeLlama-7B-Instruct & 7B & 54.0 & 38.8 & 32.9 & 50.0 & 42.3 & 45.5 & 36.6 & 48.8 & 47.2 & 50.1 & 36.9 & 40.2 & 33.2 & 42.8 \\
        DeepSeek-Coder-6.7B-Instruct & 6.7B & 74.9 & 52.2 & 30.9 & 55.9 & 64.8 & 64.7 & 25.8 & 93.8 & 59.6 & 3.3 & 65.9 & 54.8 & 47.4 & 53.4 \\
        CodeQwen1.5-7B-Chat & 7B & 77.7 & 66.6 & 66.8 & 64.4 & 66.7 & 67.5 & 67.3 & 55.1 & 60.9 & 61.1 & 65.9 & 60.0 & 54.7 & 64.2 \\
        Yi-Coder-9B-Chat & 9B & 82.0 & 73.4 & 79.1 & 70.3 & 74.1 & 73.3 & 76.4 & 90.9 & 64.4 & 60.9 & 67.3 & 63.5 & 57.3 & 71.8 \\
        Qwen2.5-Coder-14B-Instruct & 14B & \textbf{86.2} & \textbf{77.5} & \textbf{84.8} & \textbf{80.1} & \textbf{77.6} & 77.7 & \textbf{79.7} & \textbf{97.1} & \textbf{75.3} & \textbf{76.2} & \textbf{79.3} & \textbf{73.1} & \textbf{67.2} & \textbf{79.4} \\
        StarCoder2-15B-Instruct & 15B & 78.0 & 25.1 & 25.9 & 21.7 & 20.7 & 59.8 & 53.5 & 90.4 & 46.7 & 31.9 & 56.1 & 43.2 & 42.0 & 45.8 \\
        Llama-3.1-8B-Instruct & 8B & 70.1 & 59.8 & 59.1 & 56.6 & 59.1 & 59.1 & 62.5 & 85.7 & 52.2 & 42.6 & 55.9 & 44.5 & 31.8 & 56.8 \\
        OpenCoder-8B-Instruct & 8B & 79.1 & 68.1 & 71.3 & 71.0 & 67.6 & 61.4 & 68.1 & 94.4 & 66.4 & 56.1 & 70.5 & 63.1 & 56.7 & 68.8 \\
        Qwen2.5-Coder-7B-Instruct & 7B & 83.5 & 70.5 & 74.1 & 71.5 & 72.2 & 74.1 & 74.2 & 96.0 & 65.5 & 64.4 & 75.5 & 64.2 & 62.0 & 72.9 \\
        Qwen3-8B & 8B & 77.0 & 69.0 & 72.8 & 68.9 & 73.0 & 73.8 & 72.3 & 92.9 & 62.0 & 64.6 & 69.0 & 63.1 & 42.2 & 69.3 \\
        Seed-Coder-8B-Instruct & 8B & 85.2 & 72.7 & 77.0 & 74.2 & 72.8 & \textbf{78.8} & 74.7 & 95.5 & 73.4 & 72.5 & 78.0 & 70.3 & 54.2 & \textbf{75.3} \\
        \textbf{Seed-Diffusion-Preview}(0705) & - & 79.4 & 67.7 & 72.6 & 70.3 & 73.0 & 76.6 & 74.7 & 92.9 & 71.2 & 71.2 & 72.5 & 67.0 & 54.2 & 72.6 \\
        \midrule
        \multicolumn{16}{c}{\textbf{15B+ Models}} \\
        \midrule
        Codestral-22B & 22B & 78.2 & 73.6 & 77.3 & 70.1 & 71.7 & 68.5 & 74.9 & \textbf{97.1} & 71.0 & 66.6 & 74.2 & 64.4 & 50.1 & 72.1 \\
        CodeLlama-70B-Instruct & 70B & 77.8 & 66.6 & 68.6 & 69.2 & 47.8 & 62.5 & 70.5 & 77.7 & 57.2 & 51.1 & 67.0 & 51.3 & 48.7 & 62.8 \\
        DeepSeek-Coder-33B-Instruct & 33B & 80.4 & 71.8 & 76.8 & 69.9 & 72.4 & 69.8 & 75.1 & 96.4 & 70.1 & 66.6 & 75.1 & 64.6 & 54.3 & 72.6 \\
        DeepSeek-Coder-V2-Lite-Instruct & 2.4B/16B & 82.8 & 73.3 & 75.3 & 72.4 & 72.4 & 73.1 & 75.1 & 95.1 & 69.9 & 61.6 & 74.5 & 63.5 & 55.0 & 72.6 \\
        DeepSeek-Coder-V2-Instruct & 21B/236B & 89.4 & 78.2 & 77.6 & 72.6 & 74.8 & \textbf{80.5} & 75.8 & 89.1 & 74.5 & 70.7 & 80.2 & \textbf{67.9} & 59.0 & 76.2 \\
        Qwen2.5-Coder-32B-Instruct & 32B & \textbf{90.2} & \textbf{80.4} & \textbf{86.3} & \textbf{73.5} & \textbf{78.3} & 79.3 & \textbf{87.6} & 96.4 & \textbf{75.6} & \textbf{74.7} & \textbf{83.4} & 63.3 & \textbf{66.7} & \textbf{79.7} \\
        \bottomrule
    \end{tabular}
    \end{adjustbox}
\end{table}


To provide a rigorous assessment of Seed Diffusion Preview, we evaluate its performance across a diverse suite of code generation benchmarks: 

\textbf{HumanEval} and \textbf{MBPP}
We present HumanEval and MBPP results for the evaluation of basic coding ability. 

\textbf{BigCodeBench} 
BigCodeBench~\cite{zhuo2024bigcodebench} is a recent benchmark that assesses LLMs on real-world programming tasks involving multi-tool use. It features 1,140 Python tasks from 7 domains, requiring models to utilize 139 different libraries. The benchmark emphasizes compositional reasoning and is evaluated with notable rigor, using an average of 5.6 test cases and 99\% branch coverage per task. 

\textbf{LiveCodeBench}
For Competitive Coding tasks, we utilize  LiveCodeBench~\cite{jain2024livecodebench}, which continuously curates new problems from prominent competitive programming platforms including LeetCode, AtCoder and CodeForces. Crucially, it also time-stamps each problem with its release date. This temporal tagging enables the creation of contamination-free evaluation slices, ensuring models are assessed only on problems published after their training data cutoff. We provide the evaluation of all stage "v1-v6" and the most recent stage "v6" (250201-250501).

\textbf{MBXP} The MBXP benchmark~\cite{athiwaratkun2022multi} was designed for multilingual code evaluation. It adapts the problems and unit tests from the original, Python-centric MBPP benchmark for usage across more than ten programming languages.

\begin{table}[!htbp]
    \centering
    \caption{Performance on NaturalCodeBench.}
    \label{tab:naturalcodebench_performance}
    \begin{tabular}{llccccccc}
        \toprule
        \textbf{Model} & \textbf{Size} & \multicolumn{3}{c}{\textbf{NCB (zh)}} & \multicolumn{3}{c}{\textbf{NCB (en)}} & \textbf{Total} \\
        \cmidrule(lr){3-5} \cmidrule(lr){6-8}
        & & \textbf{Python} & \textbf{Java} & \textbf{Total} & \textbf{Python} & \textbf{Java} & \textbf{Total} & \\
        \midrule
        \multicolumn{9}{c}{\textbf{$\leq$15B Models}} \\
        \midrule
        CodeLlama-7B-Instruct & 7B & 18.6 & 8.6 & 13.6 & 17.1 & 14.3 & 15.7 & 14.6 \\
        DeepSeek-Coder-6.7B-Instruct & 6.7B & 38.6 & 31.4 & 35.0 & 32.9 & 32.9 & 32.9 & 33.9 \\
        Yi-Coder-9B-Chat & 9B & 41.4 & 45.7 & 43.6 & 38.6 & 44.3 & 41.5 & 42.5 \\
        Qwen2.5-Coder-14B-Instruct & 14B & 48.6 & \textbf{48.6} & 48.6 & 42.9 & 45.7 & 44.3 & 46.4 \\
        StarCoder2-15B-Instruct & 15B & 44.3 & 30.0 & 37.2 & 38.6 & 42.9 & 40.8 & 39.0 \\
        Llama-3.1-8B-Instruct & 8B & 27.1 & 24.3 & 25.7 & 22.9 & 22.9 & 22.9 & 24.3 \\
        OpenCoder-8B-Instruct & 8B & 40.0 & 30.0 & 35.0 & 35.7 & 24.3 & 30.0 & 32.5 \\
        Qwen2.5-Coder-7B-Instruct & 7B & 34.3 & 37.1 & 35.7 & 34.3 & 35.7 & 35.0 & 35.4 \\
        Qwen3-8B & 8B & 37.1 & 32.9 & 35.0 & 34.3 & 38.6 & 36.5 & 35.7 \\
        Seed-Coder-8B-Instruct & 8B & \textbf{55.7} & 45.7 & \textbf{50.7} & \textbf{50.0} & \textbf{47.1} & \textbf{48.6} & \textbf{49.6} \\
        \textbf{Seed-Diffusion-Preview}(0705) & - & 52.9 & 38.6 & 45.8 & 45.7 & 38.6 & 38.6 & 42.2\\
        \midrule
        \multicolumn{9}{c}{\textbf{15B Models}} \\
        \midrule
        Codestral-22B & 22B & 40.0 & 44.3 & 42.2 & 41.4 & \textbf{45.7} & 43.6 & 42.9 \\
        CodeLlama-70B-Instruct & 70B & 35.1 & 32.1 & 33.6 & 32.8 & 30.5 & 31.7 & 32.6 \\
        DeepSeek-Coder-33B-Instruct & 33B & \textbf{44.3} & 38.9 & 41.6 & \textbf{44.3} & 44.3 & \textbf{44.3} & \textbf{43.0} \\
        DeepSeek-Coder-V2-Lite-Instruct & 2.4B/16B & 41.4 & \textbf{47.1} & \textbf{44.3} & 41.4 & 37.1 & 39.3 & 41.8 \\
        \bottomrule
    \end{tabular}
\vspace{-10pt}
\end{table}

\textbf{NaturalCodeBench} NaturalCodeBench (NCB)~\cite{zhang2024naturalcodebench} was developed to provide a more realistic evaluation environment through a curated set of 402 problems in Python and Java, derived from genuine user queries. Its problems span across six key domains and employ complex test inputs, including varied file types and data structures.

\textbf{Aider} and \textbf{CanItEdit}
To assess code-editing capabilities, we use the Aider and CanItEdit benchmark. Aider~\footnote{https://aider.chat/docs/leaderboards/edit.html} features 133 coding exercises from Exercism, where a model must edit existing code. The primary challenge is that the model's modifications must be formatted for automated application without any human intervention. Meanwhile, the CanItEdit benchmark~\cite{cassano2023can} provides a rigorous evaluation of a model's instructional code-editing capabilities. It comprises 105 hand-crafted problems with a mix of "descriptive" (explicit) and "lazy" (ambiguous) instructions.

\subsection{Performance}
Seed-Diffusion-Preview has demonstrated huge potential of diffusion for code generation. As shown in Tables~\ref{tab:performance}, ~\ref{tab:mbxp_performance}, \ref{tab:naturalcodebench_performance} and Fig.~\ref{fig:front}, our model not only achieves performance comparable to advanced autoregressive models while operating at significantly higher speeds, but also delivers a notable boost on editing tasks. These results mark discrete diffusion as a promising direction for future exploration.

%% file: input/050conclusion.tex
\section{Discussion}
This work presents the key technical components of an experimental model from our Seed Diffusion project, demonstrating its potential for significant inference acceleration in large-scale language models. We posit that faster inference is merely the most immediate benefit of discrete diffusion. Exploring alternatives to the conventional left-to-right modeling order represents a valuable research direction, as it involves moving away from a pervasive, human-centric assumption in machine learning. Unlocking the full capabilities of discrete diffusion will require significant efforts from the community, particularly in exploring its scaling properties and its applications to complex reasoning tasks.

\newpage

\section*{Contributions}

\textbf{Project Lead}\quad 

Yuxuan Song$^{1,2,3}$, Zheng Zhang$^{1,3}$

(Alphabetical Order)

\textbf{Core Contributor}\quad

Yuxuan Song$^{1,2,3}$, Zheng Zhang$^{1,3}$, Cheng Luo$^{1}$

\textbf{Contributor}\quad

Pengyang Gao$^{1}$, Fan Xia$^{1}$, Hao Luo$^{1}$, Zheng Li$^{1}$, Yuehang Yang$^{1}$, Hongli Yu$^{1,2,3}$, Xingwei Qu$^{1}$, Yuwei Fu$^{1}$, Jing Su$^{1}$, Ge Zhang$^{1}$,  Wenhao Huang$^{1}$






\textbf{Supervision}

Mingxuan Wang$^{1,3}$, Lin Yan$^{1}$, Xiaoying Jia$^{1}$, Jingjing Liu$^{2,3}$, Wei-Ying Ma$^{2,3}$, Ya-Qin Zhang$^{2,3}$, Yonghui Wu$^{1}$, Hao Zhou$^{2,3}$

\textbf{Affiliation}





$^1$ByteDance Seed

$^2$Institute for AI Industry Research (AIR), Tsinghua University

$^3$SIA-Lab of Tsinghua AIR and ByteDance Seed

\section*{Acknowledgments}
We thank the Seed-Coder team for their help with the data pipelines and our many colleagues at ByteDance for their support of the Seed Diffusion project.




